\documentclass[letterpaper, 10 pt, conference]{ieeeconf}  
\usepackage{amsmath,amsfonts}
\usepackage{array}
\usepackage{textcomp}
\usepackage{stfloats}
\usepackage{url}
\usepackage{verbatim}
\usepackage{balance}

\usepackage{subcaption}
\usepackage{algorithm}
\usepackage{algpseudocode}
\usepackage{hhline}
\usepackage[table]{xcolor} 
\usepackage{colortbl} 
\usepackage{arydshln}
\usepackage{graphicx} 
\IEEEoverridecommandlockouts                              

\title{\LARGE \bf
	Gated Cross-Attention Network for Depth Completion
}

\author{Xiaogang  Jia, Songlei Jian, Yusong Tan, Yonggang Che, Wei Chen and Zhengfa Liang
	\thanks{Department of  Computer Science, National University of Defense Technology, China}%
	\thanks{{\tt\small  jiaxiaogang@nudt.edu.cn}}
}

\begin{document}

	\maketitle
	\thispagestyle{empty} 
	
	

	\begin{abstract}
		
	Depth completion is a popular research direction in the field of depth estimation. The fusion of color and depth features is the current critical challenge in this task, mainly due to the asymmetry between the rich scene details in color images and the sparse pixels in depth maps. To tackle this issue, we design an efficient Gated Cross-Attention Network that propagates confidence via a gating mechanism, simultaneously extracting and refining key information in both color and depth branches to achieve local spatial feature fusion. Additionally, we employ an attention network based on the Transformer in low-dimensional space to effectively fuse global features and increase the network's receptive field. With a simple yet efficient gating mechanism, our proposed method achieves fast and accurate depth completion without the need for additional branches or post-processing steps. At the same time, we use the Ray Tune mechanism with the AsyncHyperBandScheduler scheduler and the HyperOptSearch algorithm to automatically search for the optimal number of module iterations, which also allows us to achieve performance comparable to state-of-the-art methods. We conduct experiments on both indoor and outdoor scene datasets. Our fast network achieves Pareto-optimal solutions in terms of time and accuracy, and at the time of submission, our accurate network ranks first among all published papers on the KITTI official website in terms of accuracy.
		
	\end{abstract}

	\section{Introduction}

	\begin{figure*}[t]
		\begin{center}
			\includegraphics[width=\linewidth]{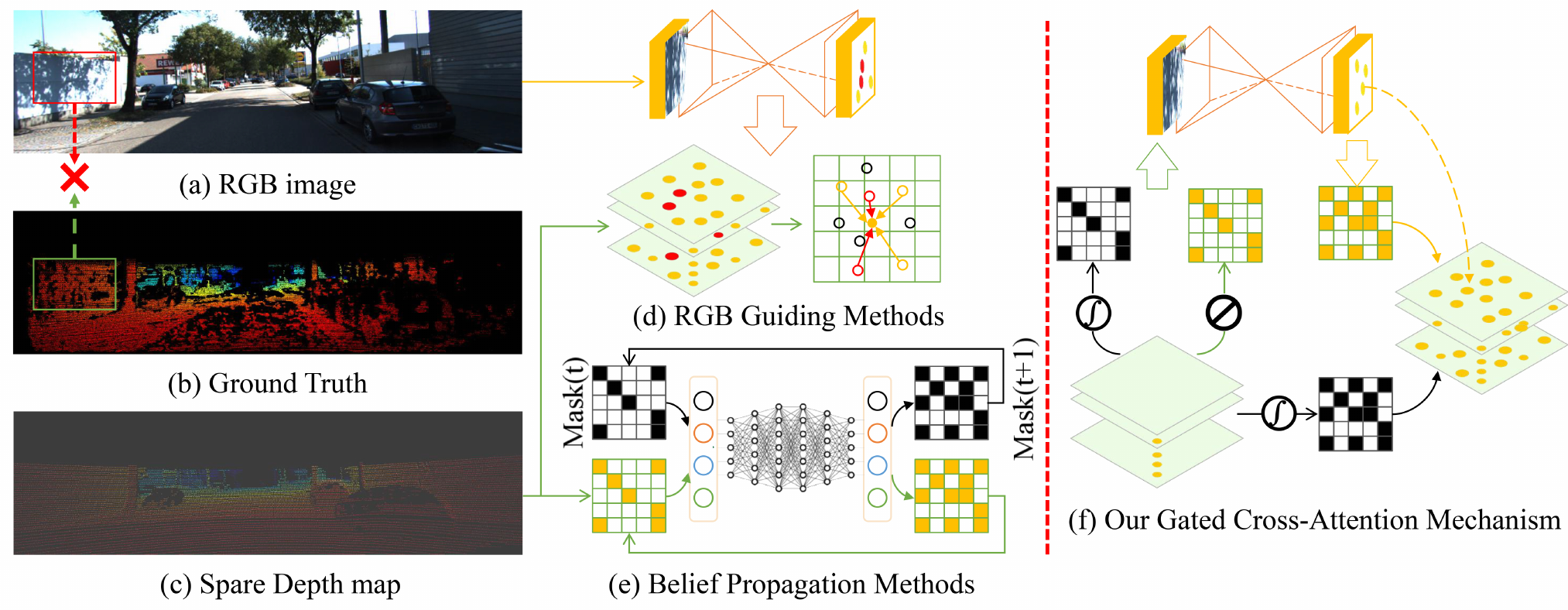}
		\end{center}
		\caption{%
			\textbf{The challenges and obstacles in depth completion.} The color image (a) contains overly complex detail features, which can cause methods that rely solely on RGB for generating guiding features to easily introduce irrelevant and incorrect information (d). The input depth map (c) is characterized by its excessively sparse features, often leading to reliance on confidence (mask) for incremental filling and optimization (e), thereby increasing the need for additional branches and computational expense. To address this, we devise a novel Gated Cross-Attention mechanism (f) that merges mask and depth information by propagating confidence features. It also employs co-attention to correct irrelevant details in the color features and to complete the depth features.
		}
		\label{fig1}
	\end{figure*}
	
	Depth estimation is critical for applications such as autonomous driving, robotic navigation, and virtual reality. However, current state-of-the-art 64-line LiDARs capture only about 5\% of effective pixels per scan, resulting in extremely sparse imaging at long distances. LiDARs with 32 and 16 lines perform even worse, failing to meet the demands of practical applications. Therefore, the task of depth completion for LiDAR-acquired maps has emerged as a significant research challenge.
	
	At present, neural network-based RGB-guided depth completion networks are the main research direction in this field, which can be categorized into early fusion \cite{dimitrievski2018learning, ma2019self, zhang2023completionformer}, late fusion \cite{tang2020learning, fan2022cascade, yan2022rignet}, spatial propagation networks \cite{cheng2018depth, cheng2020cspn++, xu2020deformable,lin2202dynamic}, and structures based on 3D representations \cite{qiu2019deeplidar, chen2019learning, xu2019depth, jeon2021abcd,liu2022nnnet,zhou2023bev}. Among these, spatial propagation networks and 3D-based structures are often considered post-processing steps, typically added to early \cite{xu2020deformable, cheng2018depth} and late \cite{hu2021penet, nazir2022semattnet} fusion frameworks to further refine the network. Early and late fusion, as mainstream frameworks in this field, focus on designing efficient feature fusion modules. Although many methods have proposed efficient module structures, such as guided feature convolution \cite{tang2020learning, wang2023decomposed}, channel shuffling \cite{liu2021fcfr, liu2023mff}, and confidence (mask) propagation \cite{eldesokey2018propagating, yan2020revisiting, schuster2021ssgp, ke2021mdanet}, they generally face the following three challenges:
	
	\subsection{Overly rich color image features}
	
	Although color images contain latent edge and detail information that can assist in the completion and optimization of depth maps, they also present a significant challenge due to the abundance of irrelevant feature information they include, as shown in Figure~\ref{fig1}. For instance, shadows and densely distributed foliage, such as leaves and shrubs, may exhibit very similar depth characteristics but display rich edge and color information in the color images. From the perspective of gradient changes, depth maps are generally smoother, whereas color images often showcase prominent color transitions. In regions where depth changes are dramatic, such as at the boundaries of occlusions between near and far scenes, the color images can appear blurry due to the necessity of central focusing during the capturing process. This can inevitably lead to the incorporation of erroneous information when relying solely on color images for feature completion, necessitating the design of more complex structures to aid the network in filtering out these irrelevant features, which in turn increases computational demand. To tackle this issue, we employ a co-attention mechanism whereby RGB features also need to be supervised and optimized by depth features. This allows for the transfer of gradient variation information from the depth maps to the RGB features and the subsequent feedback of the optimized color features into the depth features for completion. This innovative bidirectional completion approach replaces the traditional unidirectional completion mode.
	
	\subsection{Overly sparse depth features}
	
	Due to the fact that 64-line LiDAR can only capture about 5\% of pixel points, nearly all methods choose to gradually achieve depth completion through multiple rounds of optimization. However, considering that convolution operations are applied uniformly across all tensor features, this process is not friendly to discrete sparse depth features. Therefore, SparseConvs \cite{uhrig2017sparsity} introduced the concept of sparse convolution. Subsequently, researchers added a confidence (mask) branch on top of the depth branch to guide the gradual completion of depth features, as seen in methods like NConv-CNN (d) \cite{eldesokey2018propagating}, Revisiting \cite{yan2020revisiting}, SSGP \cite{schuster2021ssgp}, MDANet \cite{ke2021mdanet}, and others. However, adding extra branches at multiple scales significantly increases the computational cost. To effectively address the challenges posed by sparse depth features, we maintain normal propagation of depth features while transforming them into confidence and supervisory features through a gating mechanism. We then perform progressive optimization with completion features generated by the color branch. This approach not only achieves the propagation of confidence levels for sparse depth features but also reduces the need for additional confidence (mask) branches.
	
	\subsection{Search for Fusion Counts}
	
	As a common challenge for neural networks, hyperparameter optimization is a hurdle for all practitioners. In the field of depth completion, it requires continual experimentation to find the optimal iteration count for feature fusion modules. Previous work relied entirely on manual trial and error, generally choosing the maximum GPU memory capacity of one's own machine as the baseline. To address this challenge, we propose for the first time to search for the optimal iteration count through machine learning, significantly reducing workload and training time. Specifically, we use the Ray Tune optimizer with an AsyncHyperBandScheduler and HyperOptSearch algorithm to search for the iteration count corresponding to each scale, minimizing test loss through the scheduler to prematurely terminate unpromising experiments, thereby reducing unnecessary trials. Then, using the HyperOptSearch algorithm in conjunction with experience gained during training, we select the best combination from a tree structure composed of discrete iteration counts for experimentation, thus obtaining the optimal iteration count for the current network. The resulting network ranks first on the KITTI official website in terms of RMSE at the time of submission, with many combinations achieving competitive performance.
	
	In summary, the contributions of this paper are as follows:

	\begin{itemize}
		
		\item To address the issue of overly rich edge and detail information in color images, as well as the blurriness of distant scenes caused by central focusing, we employ a co-attention mechanism to simultaneously optimize color image features and depth features.
		
		\item In response to the challenge of excessively sparse depth features, we utilize a gating mechanism to transform them into confidence features and achieve progressive optimization, reducing the reliance on confidence (mask) branches.
		
		\item For hyperparameter optimization, we adopt the ray tune mechanism based on the AsyncHyperBandScheduler and the HyperOptSearch algorithm to search for the optimal number of module iterations, greatly reducing the trial-and-error costs of manual parameter tuning.
		
		\item Our rapid and precise depth completion methods achieve the Pareto-optimal solution, boasting the fastest speed among the top 10 ranking methods and ranking first among all methods with processing times below 100ms and 30ms. The high-precision method obtained through ray tune search attains state-of-the-art performance, with test results ranking first on the KITTI official website at the time of submission.
		
	\end{itemize}
	
	\section{Related Work}
	
	The feature fusion module is the core of late fusion in depth completion. To design an efficient module, we delve into and explore the feature fusion part of late fusion methods in the depth completion field, analyze the rationality of these modules, and thus introduce the design concept of the efficient feature fusion module proposed in this paper.
	
	\subsection{Concatenation and Channel Shuffle}
	
	Spade-RGBsD \cite{jaritz2018sparse} first proposes the strategy of late fusion, adopting concatenation in the feature fusion part and designing experiments to prove that concatenation is more effective than pixel-wise addition. Subsequently, researchers use multi-scale concatenation like PENet \cite{hu2021penet} or cross concatenation of features as in Ms\_Unc\_UARes-B \cite{zhu2022robust} to further improve accuracy. Later, FCFR-Net \cite{liu2021fcfr} employs channel shuffle extraction operations to combine more representative features from color images and rough depth maps, and uses energy-based fusion operations to select feature values with higher regional energy, proving to be advantageous over addition and simple channel concatenation. MFF-Net \cite{liu2023mff} further enhances and fuses features of different modalities by adding multi-level weighted combinations.
	
	\subsection{Confidence (Mask) Propagation}
	
	Considering the sparsity of input data, researchers define the depth completion network as a process of progressive optimization through confidence (mask) propagation. NConv-CNN (d) \cite{eldesokey2018propagating} designs an algebraically constrained convolutional layer combined with a UNet \cite{ronneberger2015u} structure to progressively determine the confidence of convolution operations, thereby achieving rapid and precise depth completion. Revisiting \cite{yan2020revisiting}, based on spatial pyramid pooling, designs a global attention module SPF and implements deep cascade feature fusion through a mask-aware module. SSGP \cite{schuster2021ssgp} proposes a sparse spatial propagation module that generates affinity blocks from RGB features and combines them with depth features and masks to achieve multi-scale feature aggregation and confidence propagation. MDANet \cite{ke2021mdanet} implements a deformable guide aggregation layer based on the deformable convolutions \cite{dai2017deformable}, with image features used to generate positional offsets and masks, guiding depth features to achieve feature aggregation as deformable convolution parameters.
	
	\begin{figure*}[t]
		\begin{center}
			\includegraphics[width=0.98\linewidth]{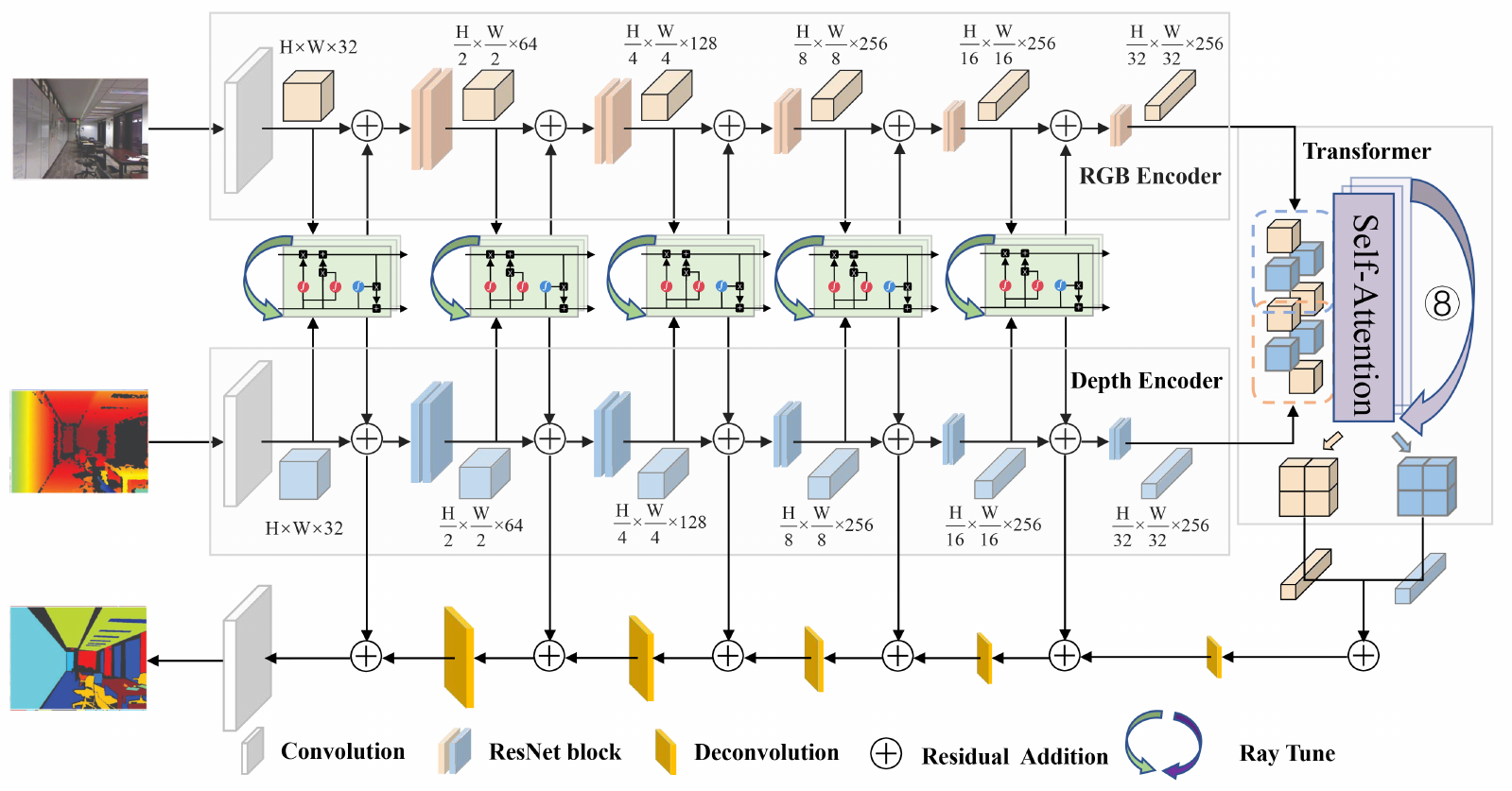}
		\end{center}
		\caption{\textbf{The overall architecture of our proposed network.} RGB images and sparse depth maps are fed into a dual-branch encoder to extract features separately. The RGB branch generates guiding features, while the depth branch produces probability features for confidence propagation. Subsequently, local pixel fusion is achieved at high resolutions through the gating mechanism, and global fusion is realized at low resolutions through the Transformer, increasing the receptive field. Features from both branches are mutually corrected and completed under the regulation of the co-attention mechanism. The number of fusion iterations is determined through progressive search using Ray Tune.  The fused features are gradually upsampled with the depth branch to the original resolution to generate the final dense and precise depth map.}
		\label{fig2}
	\end{figure*}
	
	\subsection{Attention Mechanism}
	
	Compared to direct manipulation of features or adding additional mask optimization paths, the attention mechanism can achieve feature fusion in a simple and efficient way. Channel attention mechanisms are widely used to construct fusion modules. For example, CrossGuidance \cite{lee2020deep} and SemAttNet \cite{nazir2022semattnet} add spatial attention mechanisms on top of channel attention, CDCNet \cite{fan2022cascade} includes a multi-scale pyramid structure, and RigNet++ \cite{yan2023rignet++} implements an adaptive fusion mechanism.
	
	In addition, there are methods based on guided convolution: RGB images generate guided convolution or deformable convolution parameters to optimize depth features, such as GuideNet \cite{tang2020learning}, SSGP \cite{schuster2021ssgp}, Decomposition B \cite{wang2023decomposed}, and MDANet \cite{ke2021mdanet}. However, custom convolutions lack CUDA optimization and require a significant amount of computational overhead, leading to less frequent use.
	
	Most current depth completion methods share a common understanding: RGB images generate guiding features to progressively optimize sparse depth features. However, this idea faces a core misconception: RGB images are not perfect and contain areas that require supervision, such as misaligned boundaries. To address this challenge, cross-guidance or co-attention mechanisms often achieve better results. CrossGuidance \cite{lee2020deep} proposes a cross-guidance module to refine features using attention weights obtained from its own features and those of others. DANConv \cite{yan2021dan} uses a co-attention mechanism to estimate the importance of complementary features and introduces a Symmetric Co-Attention Module (SCM) to fuse color and image domain features. ACMNet \cite{zhao2021adaptive} employs graph propagation in the encoder stage to implement a co-attention mechanism and a symmetric gating mechanism in the decoder stage to guide mutual feature fusion. BA\&GC \cite{liu2022adaptive} prevents the transmission of incorrect segmentation boundaries in RGB images into depth features and the next level of the encoder through a guided convolution module and a bidirectional attention module.
	
	Inspired by co-attention networks, we design a Gated Cross-Attention Network. The gating attention mechanism guides mutual supervision and optimization of color and depth images. The gating mechanism, combined with convolution, can achieve local pixel feature aggregation and effectively reduce computational load. At the same time, global pixel aggregation is achieved through the Transformer mechanism \cite{vaswani2017attention} at low resolution, which can effectively enhance the network's receptive field without excessively increasing computational overhead. Considering the sparsity of depth pixels and referencing the idea of confidence propagation, we use the depth branch to generate confidence features and combine them with the gating mechanism to achieve progressive optimization and sparsity propagation. In addition, we design skip connections to avoid network overfitting, emulating the design of the residual block \cite{he2016deep}.

	\section{Our Method}
	
	We design a Gated Cross-Attention Network based on the late fusion strategy and co-attention mechanism, which is primarily divided into four parts: a dual-encoder single-decoder backbone network, Gated Cross-Attention mechanism, Transformer global attention mechanism and Ray Tune iteration number optimization. Our overall network structure is shown in Figure~\ref{fig2}.
	
	\subsection{Dual-Encoder Single-Decoder Backbone Network}
	
	\algrenewcommand\algorithmicrequire{\textbf{Input:}}
	\algrenewcommand\algorithmicensure{\textbf{Output:}}
	
	\begin{algorithm}[t]
		\caption{\label{alg1}Gated Cross-Attention Mechanism}
		\begin{algorithmic}[1]
			\Require 
			\Statex Color and depth features obtained from feature extraction: $x_t$, $y_t$;
			\Ensure
			\Statex Color and depth features after fusion: $x_{t+1}$, $y_{t+1}$;\\
			Initialize: Convert depth features into confidences to determine the areas for optimization: 
			\[
			\left\{\begin{array}{l}
			f_t=x_t \times \sigma\left(W_f \otimes y_t+b_f\right)   \\
			\sigma(x)=\operatorname{sigmoid}(\mathrm{x})=\frac{1}{1+e^{-x}} 
			\end{array}\right.\]\\
			Use depth features to compute error correction vectors and optimize color features:
			\[
			\left\{\begin{array}{l}
			P_y=\sigma\left(W_p \otimes y_t+b_p\right) \\
			I_y=\tanh \left(W_i \otimes y_t+b_i\right) \\
			u_t=f_t+P_y \times I_y
			\end{array}\right.
			\] \\
			Use color features to calculate completion vectors and optimize depth features:
			\[ \left\{\begin{array}{l}
			O_y=\sigma\left(W_o \otimes y_t+b_o\right)  \\
			I_x=\tanh \left(W_t \otimes u_t+b_i\right) \\
			o_t=O_y \times I_x 
			\end{array}\right. \]\\
			Add skip-connections to transform residual vectors and enhance features:
			\[ \left\{\begin{array}{l}
			x_{t+1}=x_t+u_t  \\
			y_{t+1}=y_t+o_t 
			\end{array}\right. \]\\
			Perform progressive optimization across multiple scales and in both horizontal and vertical directions:
			\Statex \textbf{for} {$\frac{1}{2}$ \textbf{to} $\frac{1}{16}$ resolution} \textbf{do}
			\Statex \hspace{\algorithmicindent} $(x^i_{t+1}, y^i_{t+1})=(x^i_t, y^i_t) \{1 \Rightarrow 4\}$  
			\Statex \hspace{\algorithmicindent} $W, b \in \text{kernel} \{1 \times 5, 5 \times 1\}$
			\Statex \textbf{end for}
		\end{algorithmic}
	\end{algorithm}
	
	In late fusion methods, various encoder-decoder structures exist. However, to extract more refined features and reduce feature loss caused by downsampling, many methods add manually designed modules to the encoder \cite{shivakumar2019dfusenet, eldesokey2019confidence, lee2020deep, yan2020revisiting,ke2021mdanet}, redesign convolutional structures \cite{eldesokey2019confidence, yan2020revisiting, hu2021penet}, construct more complex frameworks \cite{yan2023rignet++}, or add additional branches \cite{lee2020deep,liu2022adaptive,nazir2022semattnet}. Given that the network requires multiple feature extractions, these methods all increase the computational load. At the same time, to prove the effectiveness of the core fusion module of this paper, we only use the simplest UNet \cite{ronneberger2015u} structure and widely used residual blocks \cite{he2016deep} to build the dual-branch encoder single-decoder structure.
	
	Specifically, the dual-branch encoder extracts multi-scale color and depth features, downsampling from the original resolution to $\frac{1}{32}$ resolution. The color branch generates guiding features. Considering the sparsity of the input depth map and the desire to avoid introducing additional confidence maps into the calculation, like NConv-CNN (d) \cite{eldesokey2018propagating} and SSGP \cite{schuster2021ssgp}, our depth branch directly generates confidence features. Initial feature extraction is achieved using ordinary convolution, followed by five downsampling layers to extract features from $\frac{1}{2}$ to $\frac{1}{32}$ resolution. Each downsampling layer consists of two basic residual blocks \cite{he2016deep}. During the decoder phase, features are gradually upsampled to the original resolution using deconvolution and skip connections. All convolutions are followed by Mish \cite{misra2019mish} as the activation function.
	
	\subsection{Gated Cross-Attention Mechanism}
	
	Currently, RGB images are mainly used to generate guiding features, but they include some incorrect segmentation boundaries and extraneous scene information. Depth images are sparse and require the combination with confidence (mask) maps for propagation, increasing the computational load. To address the aforementioned issues simultaneously, we design a Gated Cross-Attention mechanism, which uses the gating mechanism to achieve mutual error correction and completion between the two types of features, transforming depth features into confidence features to abandon reliance on the confidence (mask) branch, incorporating sparsity propagation within the gating mechanism.
	
	In practice, due to the sparsity of the depth map, multiple propagations are required for optimization. We place this process across multiple scales and divide the propagation into horizontal and vertical parts, propagating once per scale. Before starting propagation, we first convert depth features into confidence through  $1 \times 5$ or  $5 \times 1$ convolution layers and a sigmoid function, thereby prompting the color branch to optimize high-confidence areas. Subsequently, color features are corrected by depth features in high-confidence areas, aligning incorrect boundary areas. Finally, the corrected and aligned color features are converted and merged into the to-be-optimized high-confidence depth features, completing the depth feature completion.
	
	Considering that both the color and depth branches undergo a process of progressive optimization, we also incorporate skip connections to transform this process into residual addition, preventing gradient explosion and overfitting. This facilitates further iterations of the module or the addition of post-processing steps for more refined optimization. The overall propagation process is presented in Algorithm~\ref{alg1}:

	\subsection{Transformer Global Attention Mechanism}
	
\begin{figure}[t]
		\begin{center}
			\includegraphics[width=0.98\linewidth]{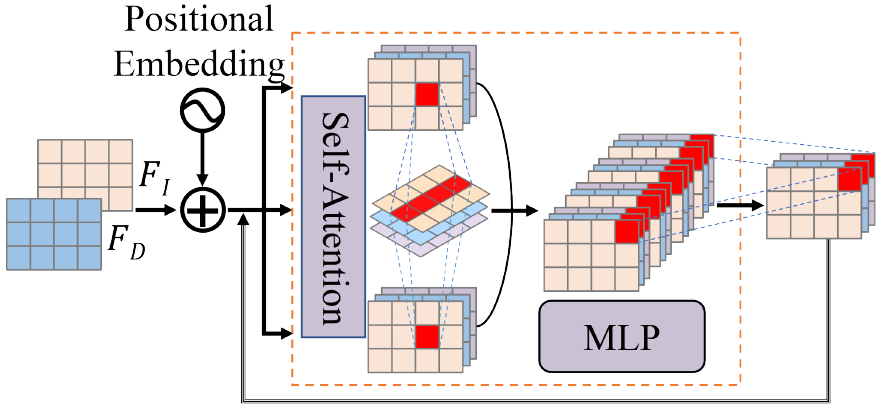}
		\end{center}
		\caption{The Transformer's global attention mechanism is implemented by executing self-attention computations at low resolution for global perception and using the multilayer perceptron to further enhance the features.}
		\label{fig3}
	\end{figure}

	The gating mechanism effectively achieves correction and completion of high-resolution local features but lacks perception of global features. Moreover, as the resolution decreases, incorrect segmentation boundaries are progressively aligned, so at low resolution, we use a Transformer variant for feature fusion and global perception. CompletionFormer \cite{zhang2023completionformer} has proven the effectiveness of the Transformer in this task, but it requires global computation at multiple scales, especially running the Transformer at high resolution, which is computationally expensive. In contrast, we only use a Transformer variant at the lowest resolution for global perception, not only greatly increasing the receptive field but also effectively reducing computational load, as shown in Figure~\ref{fig3}.
	
	Specifically, given the input features $F_I, F_D \in \mathbb{R}^{H \times W \times C}$ at $\frac{1}{32}$ resolution, where $H$, $W$, and $C$ denote the height, width, and number of channels of the feature maps, respectively, we first reduce their dimensionality to two dimensions $X_I, X_D \in \mathbb{R}^{N \times C}$, where $N$ is equal to $H \times W$. Subsequently, we concatenate the two features to form $[X_I, X_D]$ and introduce a learnable positional encoding $Y \in \mathbb{R}^{2N \times C}$. Finally, the combined features undergo global fusion and optimization through a multi-head attention mechanism.
	
	In our approach, we perform self-attention computations in a serial manner. The input for a single instance of self-attention calculation is the result from the previous self-attention, denoted as $X' \in \mathbb{R}^{2N \times C}$. Initially, we normalize it using layer normalization, and subsequently, we compute the corresponding query $\mathbf{Q}$, key $\mathbf{K}$, and value vectors $\mathbf{V}$ through linear convolution operations. Then, we carry out the self-attention operation, which can be articulated by the following formula:
	\begin{equation}
	\operatorname{Attention}(Q, K, V)=\operatorname{softmax} \left(\frac{Q \odot K^T}{\sqrt{d_K}}\right) \odot V
	\end{equation}
	where $\odot$ denotes the element-wise multiplication operation between vectors, and $d_K$ represents the number of channels in vector $\mathbf{K}$.
	
	Additionally, we enhance the features further by adding a multilayer perceptron (MLP) to the output of each self-attention layer. After the multi-head attention operation, the output is upsampled back to the input dimension. Then, building on the depth branch, we progressively upsample to the original resolution through skip connections, thereby producing the final depth result.
	
	\subsection{Iteration Count Search}
	
\begin{figure}[t]
		\begin{center}
			\includegraphics[width=0.98\linewidth]{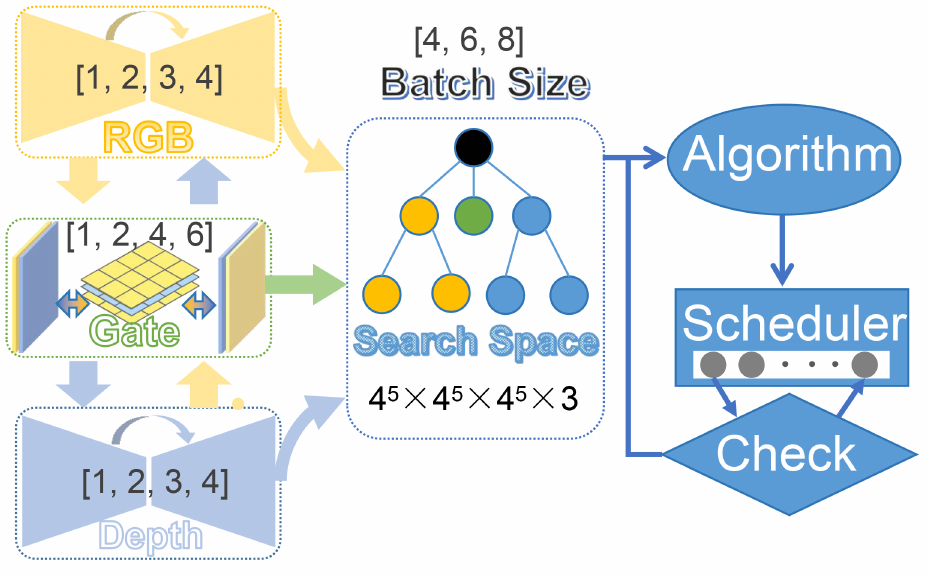}
		\end{center}
		\caption{\textbf{Schematic diagram of the iteration count search strategy.} We employ automated machine learning to find the optimal solution from the vast search space.}
		\label{fig4}
	\end{figure}

	Although our method can achieve fast and accurate results with a single iteration, it still falls short compared to state-of-the-art methods. Moreover, many methods \cite{tang2020learning, yan2023rignet++,cheng2020cspn++,zhang2023completionformer} have proven that performing multiple fusions at a single scale can significantly enhance performance. However, they typically set the number of iterations based on experience (usually a multiple of 2 and as high as the machine's maximum GPU memory capacity allows). Manually setting the number of iterations is highly limited, as the search space for feature extraction and fusion across multiple scales can be as large as \(k_i^{3  \times \text{scale}}\) (where \(k_i\) is the number of feature extractions or fusions at scale \(i\)). With a minimum \(k_i\) of 2 and scale of 5, the search space includes 32768 possibilities, making it extremely difficult to determine the optimal combination through just a few experiments. To address this, we automate the search for the best iteration count using machine learning, as shown in Figure~\ref{fig4}. The HyperOptSearch algorithm selects more promising combinations from a tree-structured discrete search space to add to the experiment queue, while the AsyncHyperBandScheduler supervises training to terminate poorly performing experimental combinations early. The intermediate results obtained from experiments can also serve as prior knowledge to help improve the HyperOptSearch algorithm, thereby minimizing test loss to achieve the best hyperparameter combination.
	
	Our hyperparameter search is highly suitable for the field of depth completion, particularly for late fusion methods. It moves beyond the previous reliance on manual trial-and-error and enables the search space to be freely expanded. Our experiments also reveal that a greater number of iterations does not always equate to better results. The optimal solution often lies in ingenious combinations.

	\subsection{Loss function}
	
	We employ a weighted sum of mean squared error (MSE) and root mean squared error (RMSE) as the loss function, which not only allows for rapid convergence to the minimum, thereby shortening the training time, but also yields a superior RMSE model, thereby enhancing the testing performance.
	\begin{equation}
	\begin{split}
	L = & \alpha \cdot \left(\frac{1}{N^i} \sum_i \operatorname{Bool}\left(d_{gt}^i>e^{-3}\right)\left(d_{pre}^i-d_{gt}^i\right)^2\right) \\
	& + \beta \cdot \sqrt{\frac{1}{N^i} \sum_i \operatorname{Bool}\left(d_{gt}^i>e^{-3}\right)\left(d_{pre}^i-d_{gt}^i\right)^2}
	\end{split}
	\end{equation}
	where \(d_{pre}\) denotes the dense depth values obtained by the decoder, \(d_{gt}\) represents the ground truth values, and \(\operatorname{Bool}\left(d_{gt}^i > e^{-3}\right)\) signifies that only the pixels in the ground truth with values greater than 0.001 are considered for comparison and computation. The coefficients \(\alpha\) and \(\beta\) are utilized to control the training stage, implementing the process from pre-training to fine-tuning.
	
	\section{Experiments}
	
	We construct the network architecture using PyTorch and conduct experiments on NVIDIA A100 GPUs. We also employ automated hyperparameter tuning tools to assist in training the network and perform comparisons on mainstream public datasets. In addition, we validate the effectiveness of each module through ablation studies. The specific experimental details are as follows:
	
	\begin{table*}[t]
		\caption{\label{tab1}\textbf{Quantitative test results for indoor and outdoor datasets.} We conduct comparisons between methods that prioritize high precision and those that are both fast and accurate. The evaluation metrics for each method are derived from their respective published papers. Our proposed methods can achieve both rapid and precise results as well as state-of-the-art performance.}
		\centering
		\begin{tabular}{l|cccc|cll|ccc}
			\hhline{------~----}
			\textit{\bfseries KITTI DC}         & \cellcolor{blue!25} \begin{tabular}[c]{@{}c@{}}RMSE\\ (mm) $\downarrow$\end{tabular} & \begin{tabular}[c]{@{}c@{}}MAE\\ (mm) $\downarrow$\end{tabular} & \begin{tabular}[c]{@{}c@{}}iRMSE\\ (1/km) $\downarrow$\end{tabular} & \begin{tabular}[c]{@{}c@{}}iMAE\\ (1/km) $\downarrow$\end{tabular} & \cellcolor{blue!25}\begin{tabular}[c]{@{}c@{}}Time\\ (s) $\downarrow$\end{tabular} &  & \textit{\bfseries NYUv2}         & \cellcolor{blue!25}\begin{tabular}[c]{@{}c@{}}RMSE\\ (m) $\downarrow$\end{tabular} & \begin{tabular}[c]{@{}l@{}}REL\\ (m) $\downarrow$\end{tabular} & \begin{tabular}[c]{@{}c@{}}$\delta_{\substack{1.25 \\ \uparrow}}$\end{tabular} \\ \hhline{======~====}
			GuideNet \cite{tang2020learning}        & 736.24 & 218.83 & 2.25 & 0.99 & 0.14  &  & S2D \cite{ma2019self}          & 0.230 & 0.044 & 97.1 \\
			CompletionFormer \cite{zhang2023completionformer} & 708.87 & 203.45 & 2.01 & 0.88 & 0.12  &  & CSPN \cite{cheng2018depth}    & 0.117 & 0.016 & 99.2 \\
			Decomposition \cite{wang2023decomposed}   & 707.93 & 205.11 & 2.05 & 0.91 & 0.1  &  & DeepLiDAR \cite{qiu2019deeplidar}    & 0.115 & 0.022 & 99.3 \\
			BEV@DC \cite{zhou2023bev}         & 697.44 & 189.44 & 1.83 & 0.82 & 0.1  &  & ACMNet \cite{zhao2021adaptive}      & 0.105 & 0.015 & 99.4 \\
			LRRU \cite{wang2023lrru}            & 695.67 & 198.31 & 2.18 & 0.86 & 0.12  &  & NLSPN \cite{park2020non}        & 0.092 & 0.012 & 99.6 \\
			\cdashline{1-6}[0.8pt/2pt]
			Ours-accurate   & 707.53 & 213.04 & 2.14 & 0.97 & 0.047 &  & GuideNet \cite{tang2020learning}     & 0.101 & 0.015 & 99.5 \\
			\cline{1-6}
			ENet \cite{hu2021penet}            & 741.30 & 216.26 & 2.14 & 0.95 & 0.019  &  & FCFR-Net \cite{liu2021fcfr}     & 0.106 & 0.015 & 99.5 \\ 
			MDANet \cite{ke2021mdanet}         & 738.23 & 214.99 & 2.12 & 0.99 & 0.03  &  & RigNet \cite{yan2022rignet}      & 0.090 & 0.012 & 99.6 \\ 
			SPL \cite{liang2022selective}             & 733.44 & 212.49 & 2.09 & 0.93 & 0.03   &  & DySPN \cite{lin2202dynamic}        & 0.090 & 0.012 & 99.6 \\ 
			ReDC \cite{sun2023revisiting}            & 728.31 & 204.60 & 2.05 & 0.89 & 0.02   &  & Decomposition \cite{wang2023decomposed}& 0.098 & 0.014 & 99.5 \\ 
			NNNet \cite{liu2022nnnet}          & 724.14 & 205.57 & 1.99 & 0.88 & 0.03  &  & BEV@DC \cite{zhou2023bev}      & 0.089 & 0.012 & 99.6 \\
			\cdashline{1-6}[0.8pt/2pt] 
			Ours-fast       & 720.98 & 218.96 & 2.31 & 1.02 & 0.023 &  & LRRU \cite{wang2023lrru}        & 0.091 & 0.011 & 99.6        \\ 
			\cdashline{8-11}[0.8pt/2pt]
			Ours-middle       & 717.71 & 213.11 & 2.31 & 0.99 & 0.027 &  & Ours-accurate& 0.090 & 0.012 & 99.6        \\ 
			\cline{1-6} \cline{8-11} 
		\end{tabular}
	\end{table*}

	\begin{figure*}
		\centering
		\begin{subfigure}{0.23\linewidth}
			\includegraphics[width=1\textwidth]{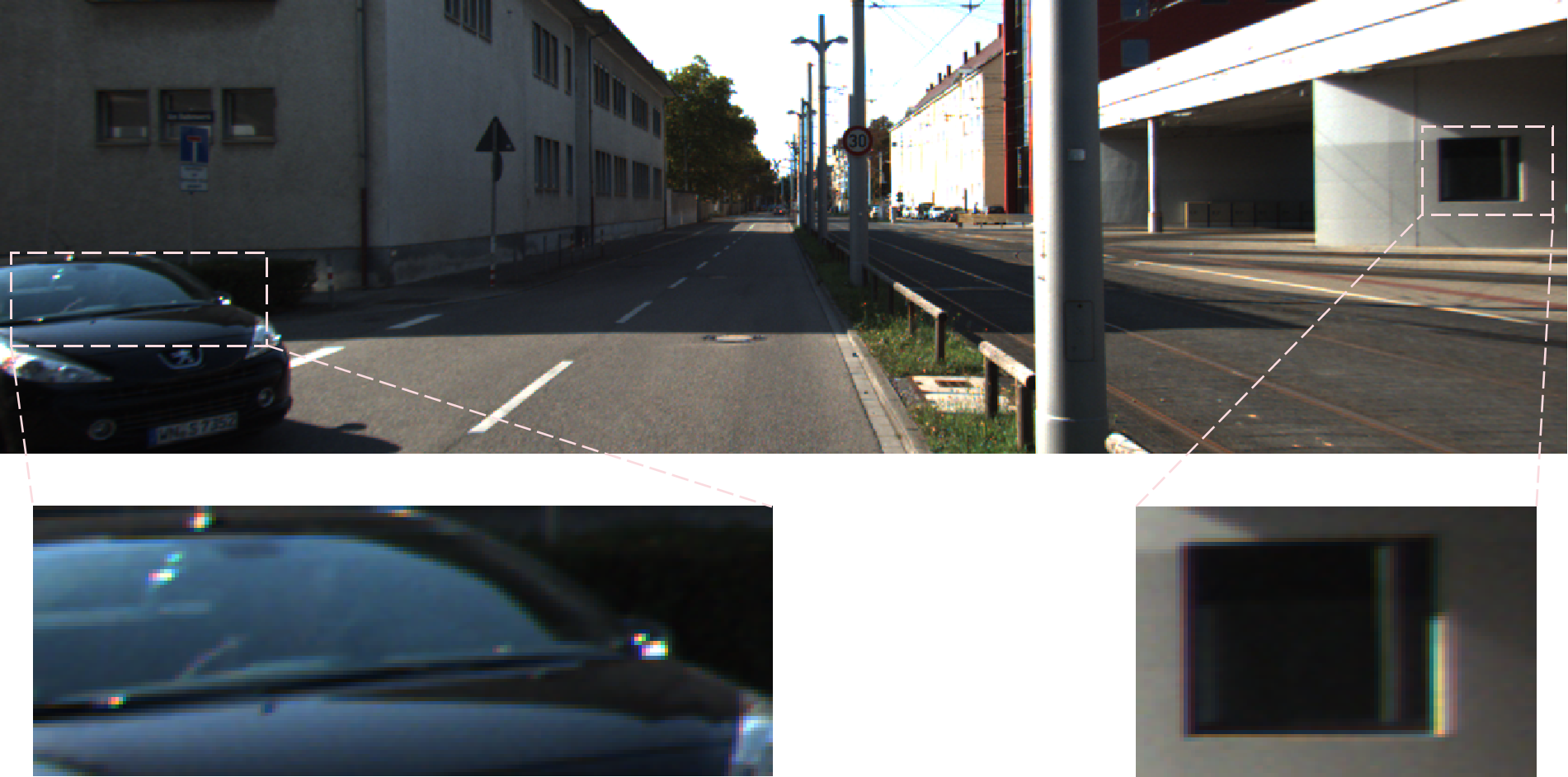}
			\caption{RGB image}
		\end{subfigure}
		\hfill
		\begin{subfigure}{0.23\linewidth}
			\includegraphics[width=1\textwidth]{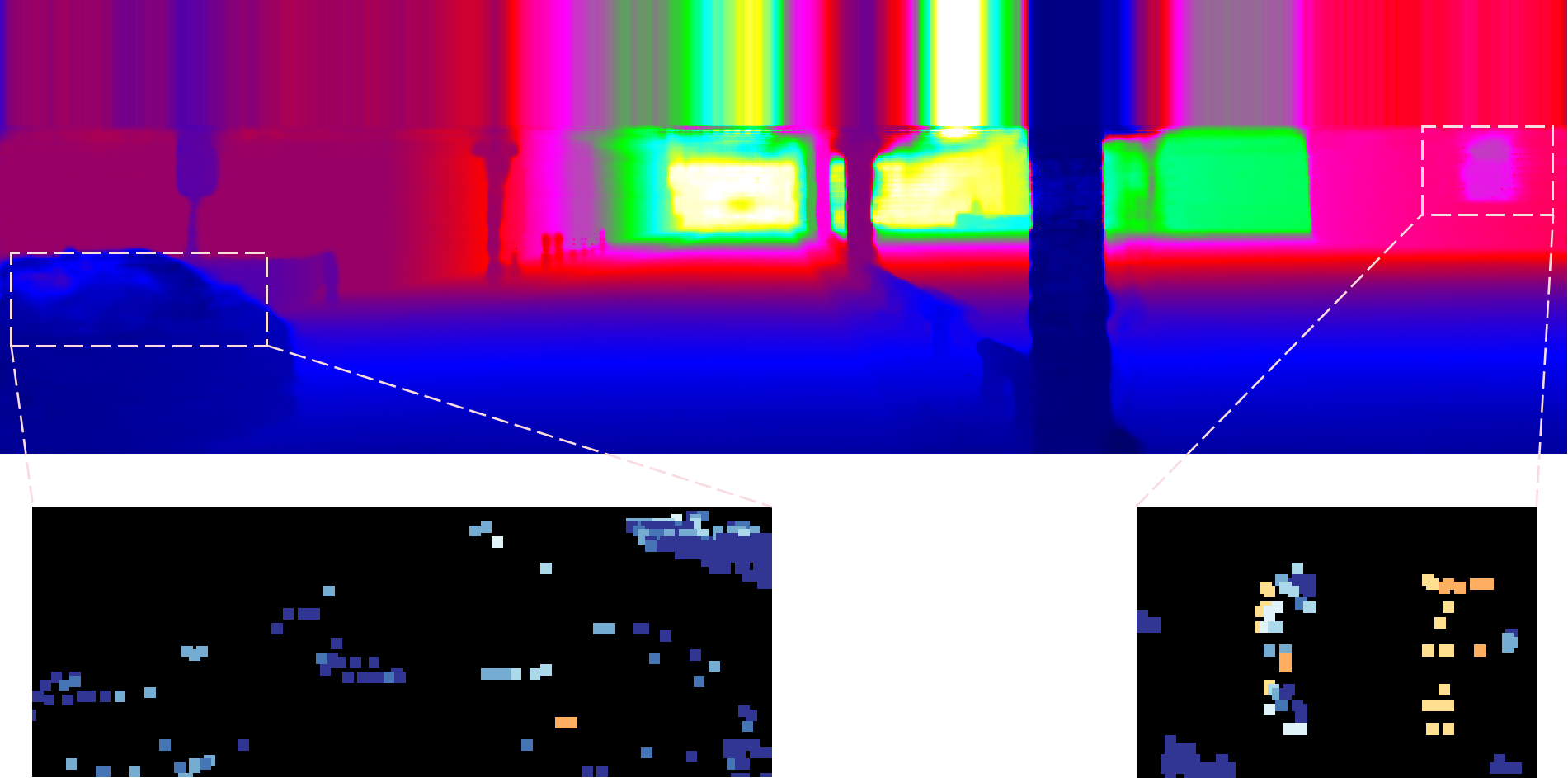} 
			\caption{GuideNet}
		\end{subfigure}
		\hfill
		\begin{subfigure}{0.23\linewidth}
			\includegraphics[width=1\textwidth]{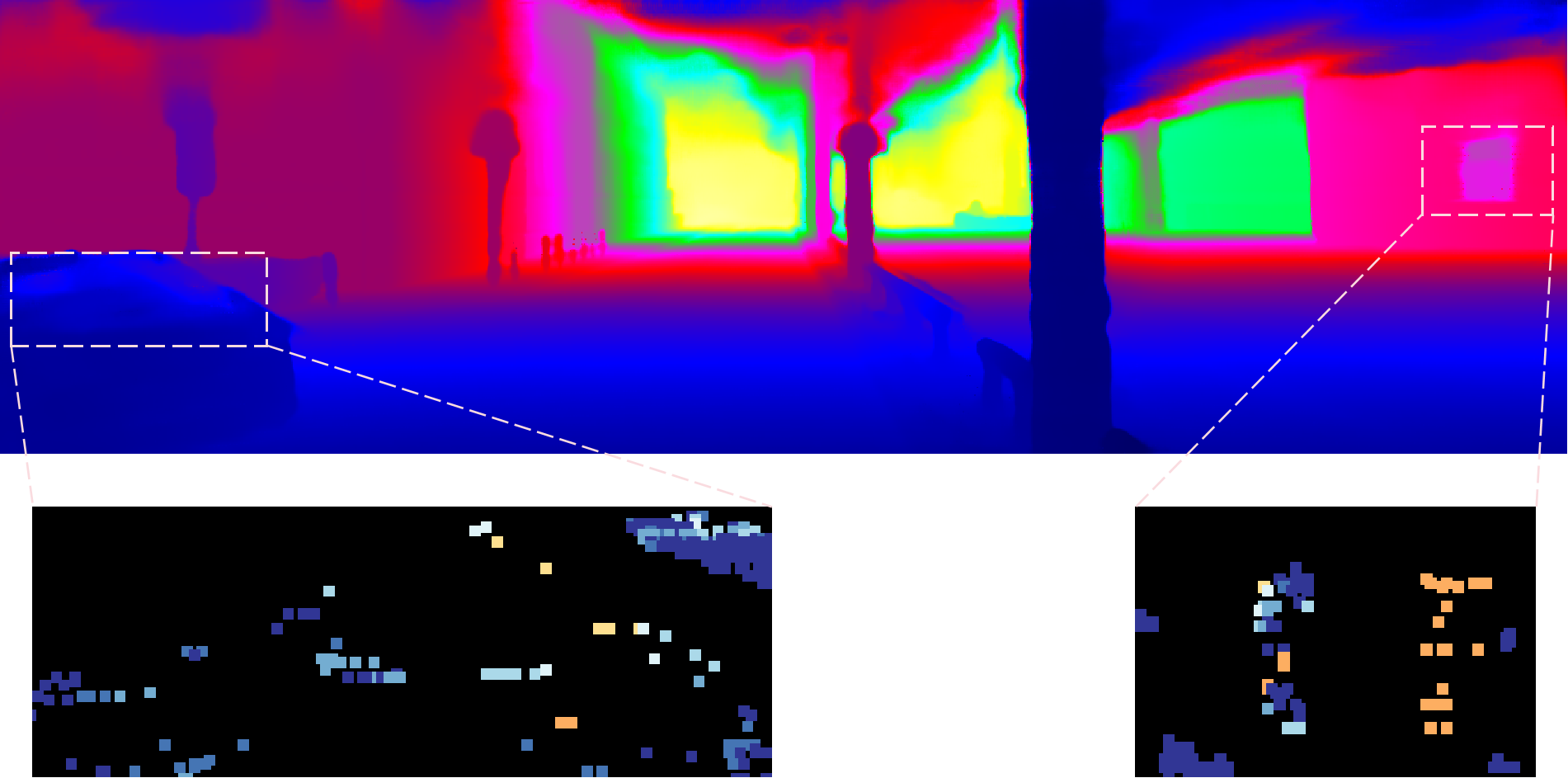}
			\caption{DySPN}
		\end{subfigure}
		\hfill
		\begin{subfigure}{0.23\linewidth}
			\includegraphics[width=1\textwidth]{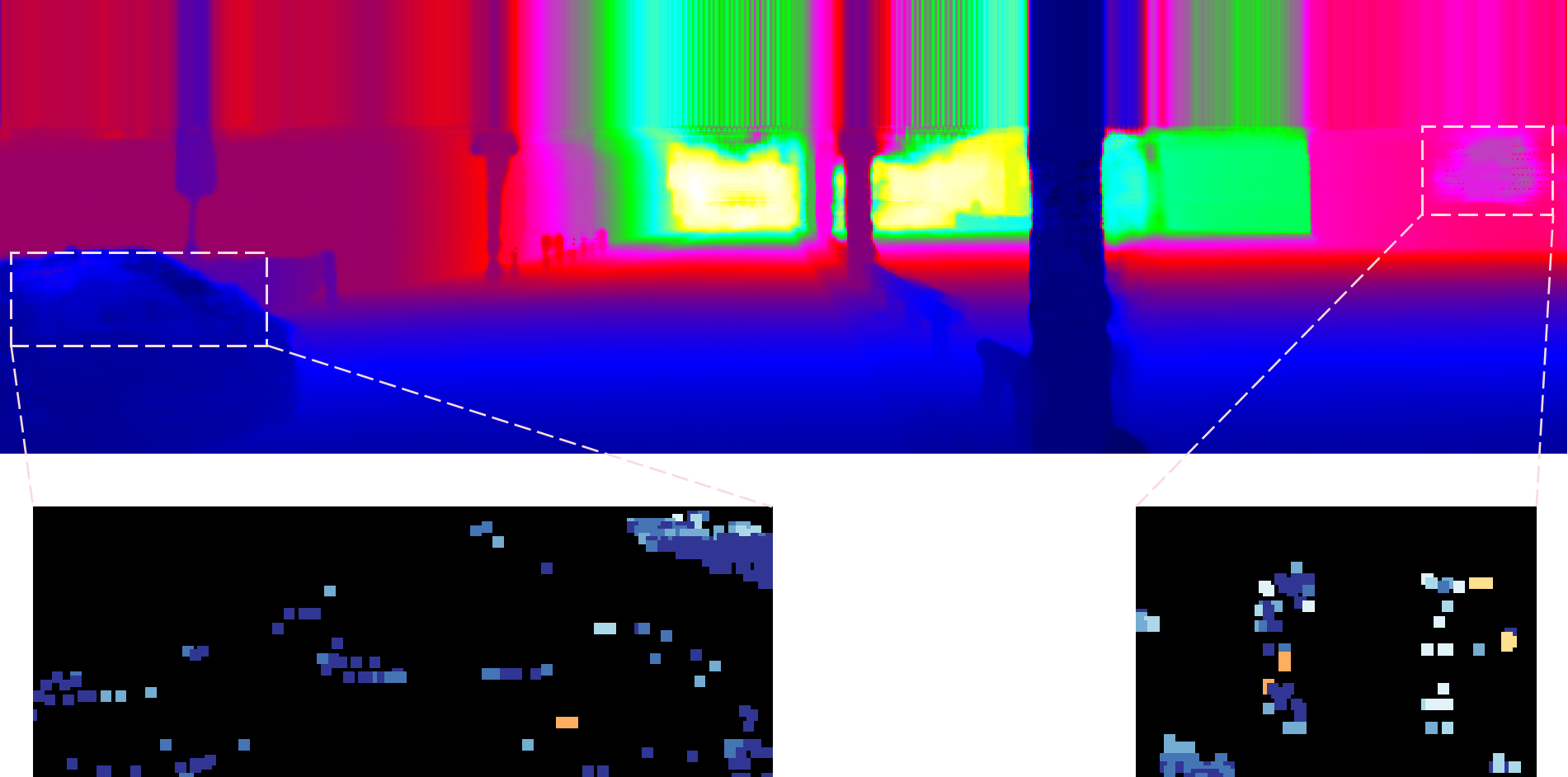}
			\caption{Ours}
		\end{subfigure}
		
		\begin{subfigure}{0.23\linewidth}
			\includegraphics[width=1\textwidth]{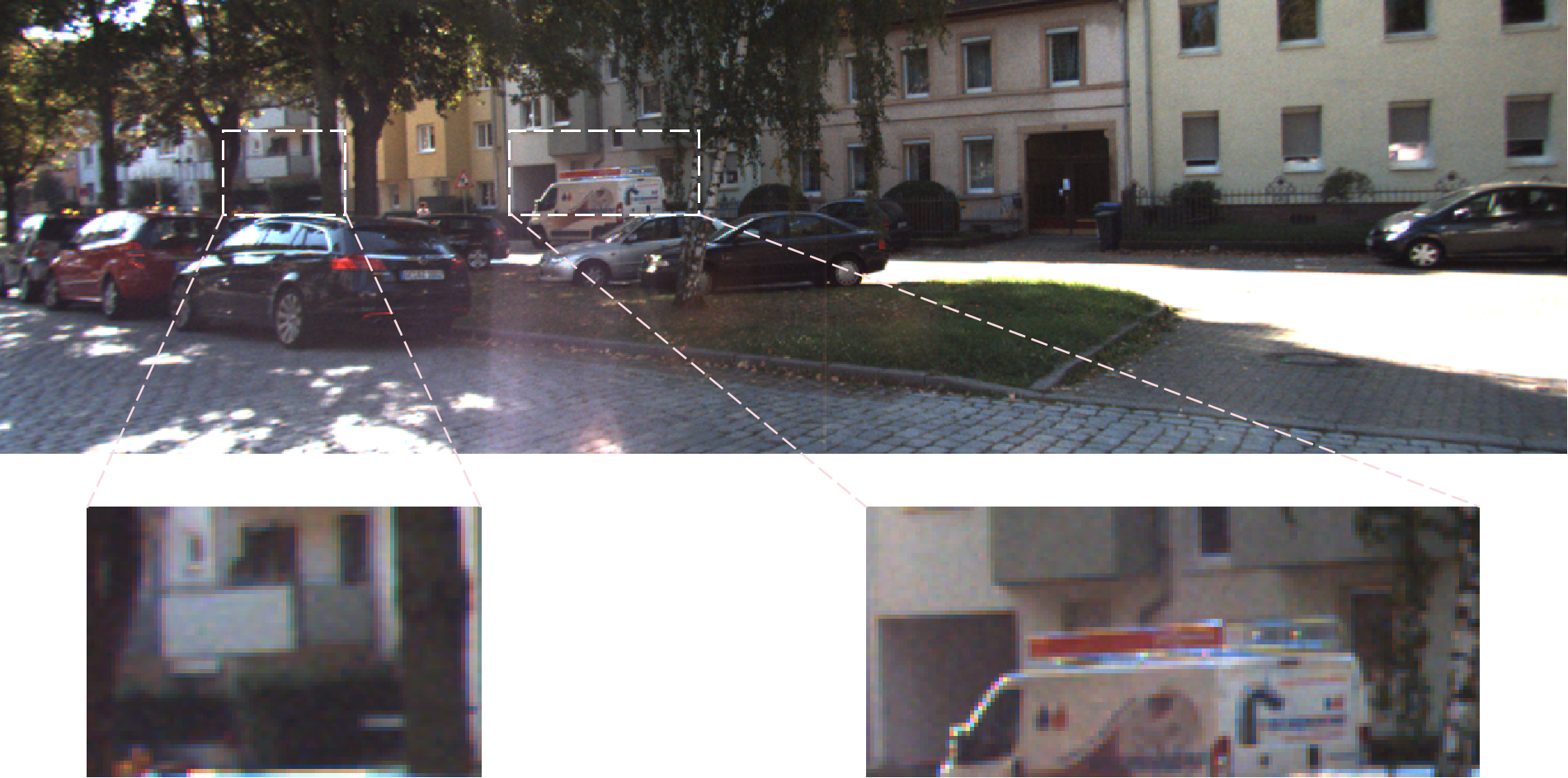}
			\caption{RGB image}
		\end{subfigure}
		\hfill
		\begin{subfigure}{0.23\linewidth}
			\includegraphics[width=1\textwidth]{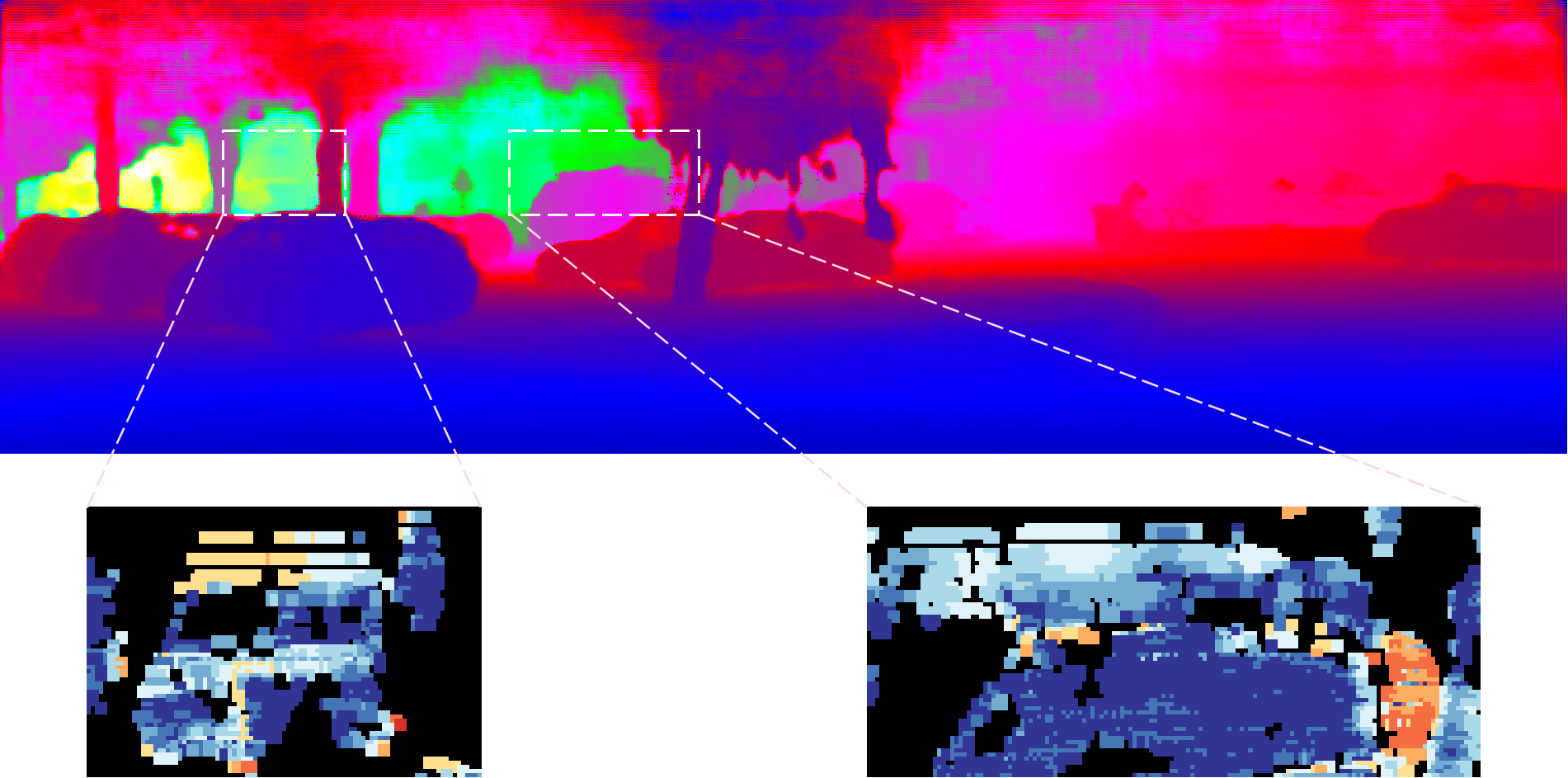}
			\caption{ENet}
		\end{subfigure}
		\hfill
		\begin{subfigure}{0.23\linewidth}
			\includegraphics[width=1\textwidth]{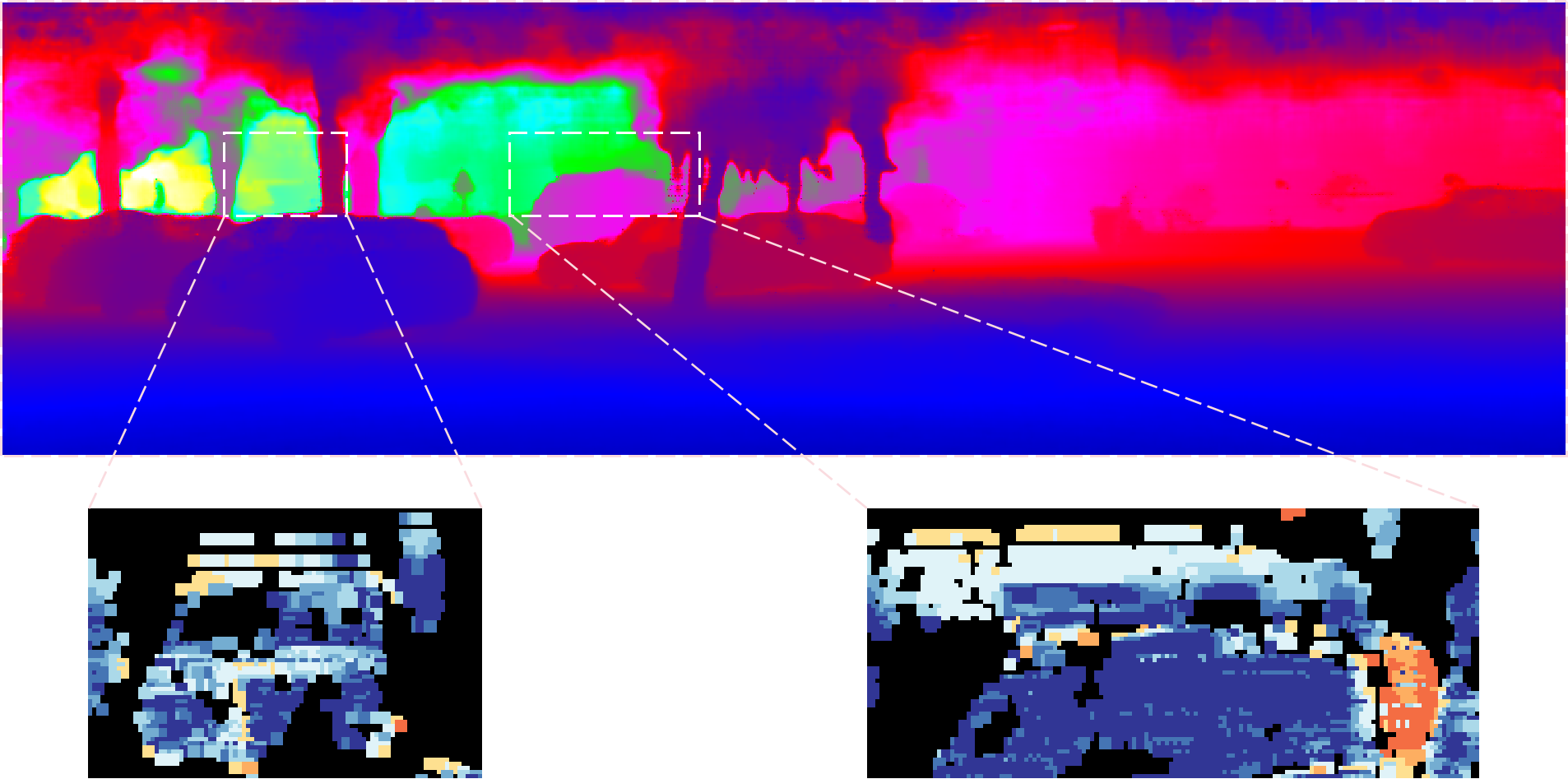}
			\caption{MSG-CHN}
		\end{subfigure}
		\hfill
		\begin{subfigure}{0.23\linewidth}
			\includegraphics[width=1\textwidth]{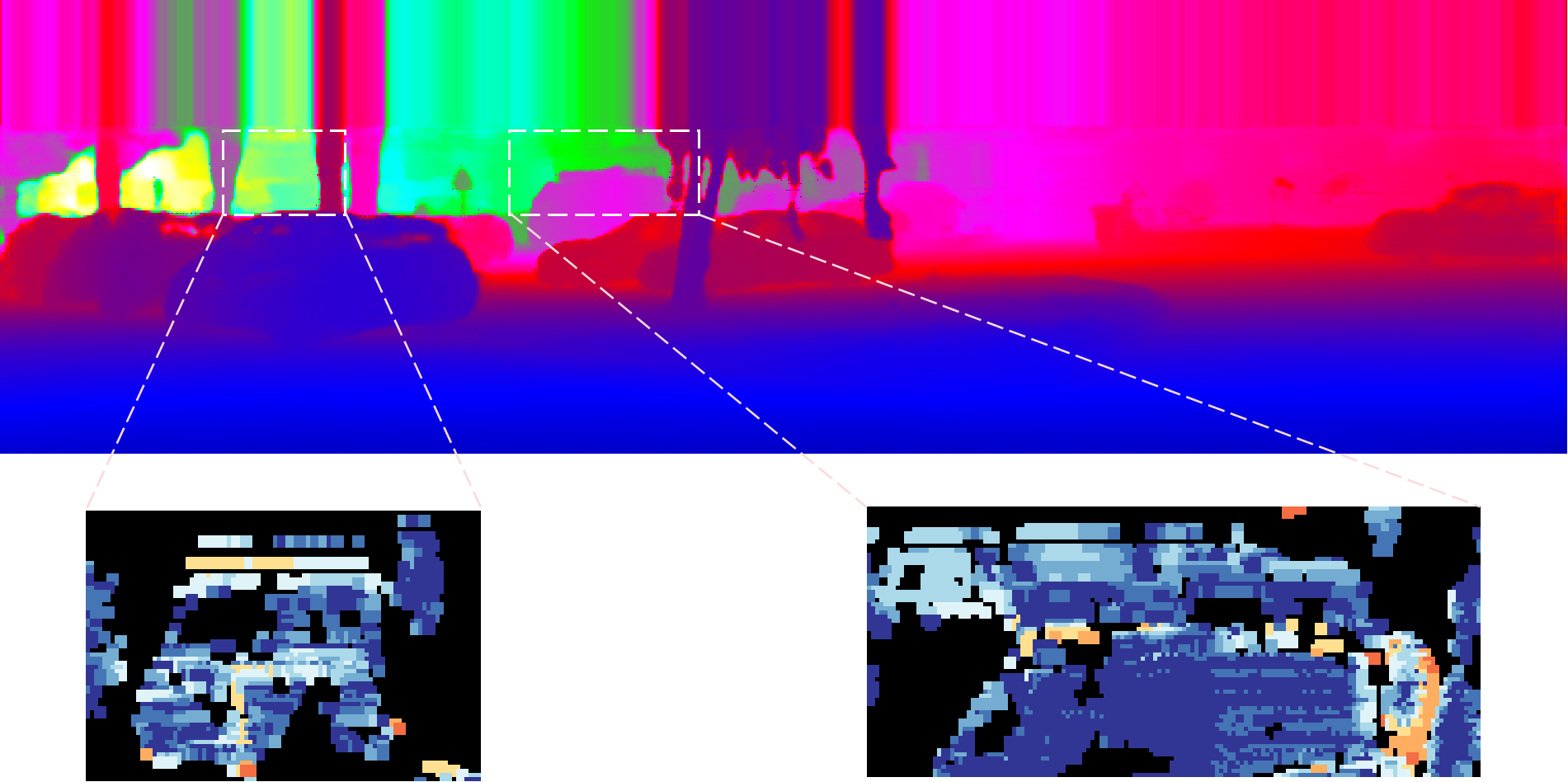}
			\caption{Ours}
		\end{subfigure}
		\caption{\textbf{Comparative visualization results for official test images from the KITTI Depth Completion benchmark.} We select some of the latest representative methods for comparison, including Baseline, Late Fusion, and Spatial Propagation Networks, etc.}
		\label{fig5}
	\end{figure*}

	\subsection{Datasets and Evaluation Metrics}

	\textbf{KITTI Depth Completion Benchmark \cite{Uhrig2017THREEDV}:}
	The KITTI Depth Completion Benchmark (KITTI DC) is the current mainstream dataset and testing benchmark for depth completion, and it is also a large-scale real-world autonomous driving dataset. This dataset is collected by the HDL-64 LiDAR. The sparse depth maps correspond to the original LiDAR scan points under the RGB camera view, containing less than 5\% of valid pixel values. The dense depth maps (ground truth) are obtained by accumulating LiDAR scan points over multiple timestamps and projecting them together, and then removing occlusions and moving objects' abnormal depth values through stereo depth comparison of image pairs, which contain about 14\% of valid pixels. The official split includes 86,898 images for training, 1,000 images for validation, and an additional 1,000 images without dense depth maps for testing. A website is provided for submitting 1,000 test results for comparison with other depth completion methods. Evaluation metrics include Root Mean Squared Error (RMSE), Mean Absolute Error (MAE), Root Mean Squared Error of the Inverse Depth (iRMSE), and Mean Absolute Error of the Inverse Depth (iMAE). The specific values are calculated and displayed on the official website, and the calculation formulas are as follows:
	
	\begin{itemize}
		\item RMSE (mm): $\sqrt{\frac{1}{N^i} \sum_{i} \left(d_{pre}^i - d_{gt}^i\right)^2}$
		\item MAE (mm): $\frac{1}{N^i} \sum_{i} \left|d_{pre}^i - d_{gt}^i\right|$
		\item iRMSE (1/km): $\sqrt{\frac{1}{N^i} \sum_{i} \left|\frac{1}{d_{pre}^i} - \frac{1}{d_{gt}^i}\right|^2}$
		\item iMAE (1/km): $\frac{1}{N^i} \sum_{i} \left|\frac{1}{d_{pre}^i} - \frac{1}{d_{gt}^i}\right|$
	\end{itemize}
	
	\textbf{NYU Depth Dataset V2 \cite{silberman2012indoor}:}
	The NYU Depth Dataset V2 is the current mainstream indoor scene dataset for depth completion. This dataset consists of RGB and depth images captured by the Microsoft Kinect camera from 464 indoor scenes, with an image resolution of $480 \times 640$. We use the official split of 50,000 images for training and test on the 654 marked images. Evaluation metrics are similar to other methods, including Root Mean Squared Error (RMSE), Relative Error (REL), and Pixel Ratio, with the calculation methods of each metric as follows:
	
	\begin{itemize}
		\item Relative Error (REL): $\frac{1}{N^i} \sum_{i} \left|\frac{d_{pre}^i - d_{gt}^i}{d_{gt}^i}\right|$
		\item Threshold Accuracy ($\delta_\tau$): $\left[  \frac{1}{N^i} \sum_{i} \max \left(\frac{d_{gt}^i}{d_{pre}^i}, \frac{d_{pre}^i}{d_{gt}^i}\right)\right] < \tau , \tau \in \left\{1.25\right\}$
	\end{itemize}
	
	\subsection{Quantitative Analysis}

	We utilize Ray Tune for hyperparameter tuning and training, selecting the best-performing experimental results from a total of 50 trials. Each trial is planned to train for 30 epochs with an initial learning rate set at 0.001 and weight decay at 0.05. The learning rate is halved at the 10th, 20th, and 25th epochs, and the loss function is fine-tuned during the last 5 epochs. At the 10th epoch, we conduct a test every 500 batches to save intermediate model results. Additionally, the scheduler sets a checkpoint every 5 epochs after reaching the 10th epoch, eliminating a quarter of the experiments with poorer performance by minimizing the RMSE value. Following this, we employ the HyperOptSearch algorithm in conjunction with the scheduler's outcomes to identify more promising hyperparameter combinations for new trials. This process refines the hyperparameter tuning and progressively yields the best combinations and intermediate training models. We then evaluate these models using test data, calculate other performance metrics, and compare them as shown in Table~\ref{tab1}.

	\begin{table*}[]
		\caption{\label{tab2}\textbf{Ablation study results on the KITTI official selval dataset.} "Confidence", "Depth-\textgreater{}RGB", and "RGB-\textgreater{}Depth" respectively represent the confidence conversion in the gating mechanism, depth features guiding color features, and color features completing depth features. "Transformer" denotes the low-resolution global perception module, and Ray Tune is used to search for the optimal number of iterations.}
		\centering
		\begin{tabular}{c|ccccc|c}
			\hhline{-------}
			\textit{\textbf{KITTI DC}} & Confidence & Depth-\textgreater{}RGB & RGB-\textgreater{}Depth  & Transformer  & Ray Tune  & \cellcolor{blue!25}\begin{tabular}[c]{@{}c@{}}RMSE\\ (mm)  $\downarrow$ \end{tabular}  \\ \hhline{=======}
			(a)                        & $\times$  & \checkmark               & \checkmark               & \checkmark   & $\times$ & 810.2                                                                                          \\
			(b)                        & \checkmark  & $\times$               & \checkmark               & \checkmark   & $\times$ & 784.1                                                                                           \\
			(c)                        & \checkmark  & \checkmark               & $\times$               & \checkmark   & $\times$ & 776.3                                                                                          \\
			(d)                        & \checkmark  & \checkmark               & \checkmark               & $\times$   & $\times$ & 773.4                                                                                            \\
			(e)                        & \checkmark  & \checkmark               & \checkmark               & \checkmark   & $\times$ & 762.1                                                                  \\ \cdashline{1-7}[0.8pt/2pt]
			(f)                        & \checkmark  & \checkmark               & \checkmark               & \checkmark   & middle       & 754.1                                                                                           \\
			(g)                        & \checkmark  & \checkmark               & \checkmark               & \checkmark   & accurate   & 751.7                                                                                            \\ 
			\hline
		\end{tabular}
	\end{table*}
	
	\begin{table*}[t]
		\caption{\label{tab3}\textbf{Official KITTI benchmark evaluation results before and after the integration of SPNs.} We conduct tests on methods that prioritize speed and high precision and find that results obtained through hyperparameter tuning significantly outperform those with added SPNs. Moreover, spatial propagation networks show a noticeable loss in accuracy when applied to larger models.}
		\centering
		\begin{tabular}{l|cccc|c}
			\hline
			\textit{\bfseries KITTI DC} & \cellcolor{blue!25}RMSE (mm) $\downarrow$ & MAE (mm) $\downarrow$ & iRMSE (1/km) $\downarrow$ & iMAE (1/km) $\downarrow$ & \cellcolor{blue!25}Time (s) $\downarrow$ \\
			\hline
			Ours-fast         & 720.98 & 218.96 & 2.31 & 1.02 & 0.023 \\
			Ours-fast+CSPN++ & 711.08 & 204.44 & 2.10 & 0.90 & 0.086 \\
			Ours-fast+NLSPN    & 720.42 & 210.69 & 2.15 & 0.93 & 0.044 \\
			\hline
			Ours-middle          & 717.71 & 213.11 & 2.31 & 0.99 & 0.027 \\
			\hline
			Ours-accurate           & 707.53 & 213.04 & 2.14 & 0.97 & 0.047 \\
			Ours-accurate+CSPN++       & 714.47 & 206.97 & 2.08 & 0.90 & 0.105 \\
			Ours-accurate+NLSPN      & 885.28 & 259.49 & 3.18 & 1.21 & 0.088 \\
			\hline
		\end{tabular}
	\end{table*}

	From the table, it is clear that our method not only achieves rapid and precise depth completion results but also shows significant improvement compared to the baseline (GuideNet \cite{tang2020learning}). Furthermore, the optimal iteration model identified through hyperparameter tuning search is capable of reaching state-of-the-art performance. In addition, our proposed fast method ranks first among all methods with a processing time below 100ms and 30ms, and our high-precision method ranks first among all published papers when submitted to the KITTI official website.

	\subsection{Qualitative Analysis}
	
	We select a subset of the test results from the official website for visual analysis, as depicted in Figure~\ref{fig5}. Our method successfully generates accurate depth values for the transition pixels between distant and nearby areas, which tend to blur due to the central focus of the RGB camera. By using a cross-attention approach that leverages depth features to enhance the color characteristics of blurred regions, we effectively tackle the optimization challenges for pixels in these areas.

	\subsection{Ablation study}
	
	To verify the importance of each module to the network, we conduct an analysis on all elements, including the gating mechanism, Transformer, and hyperparameter optimization. This serves to confirm the influence of confidence propagation, cross-attention, global perception, and hyperparameter optimization on the depth completion results, as demonstrated in Table~\ref{tab2}.

	As shown in Table~\ref{tab2}, the comparison between (a) and (e) reveals that omitting the confidence conversion process results in a loss of accuracy. This is due to the sparsity of the input depth, which relies on and benefits from confidence propagation. The contrast between (b) and (e) shows that adding the supervision of RGB features by depth features significantly increases accuracy, as the color images also require supervision from depth features, confirming the importance of bidirectional supervision as evidenced by (c). The comparison between (d) and (e) indicates that global perception is highly effective in improving accuracy because the self-attention mechanism greatly expands the network's receptive field. Comparing (e), (f), and (g), it is apparent that searching for the number of iterations with Ray Tune noticeably enhances accuracy without adding much time cost (0.004s and 0.024s), and the improvement in accuracy relies on a clever combination of hyperparameters rather than simply increasing the number of modules.
	
	\subsection{Adding SPN}
	
	To validate the generalizability of the proposed method and further enhance accuracy, we add a spatial propagation network post-processing step on top of our methods, including CSPN++ \cite{cheng2020cspn++}, NLSPN \cite{park2020non}. As shown in Table~\ref{tab3}, adding SPN to the fast method slightly improves the accuracy (0.56, 9.9), but it comes with a significant time cost (21ms, 63ms) due to the need for full-resolution feature computation in spatial propagation networks. This presents a considerable risk for real-time applications such as autonomous driving. In contrast, by using Ray Tune to search for the optimal hyperparameter combinations, we can significantly boost accuracy (13.45) without excessive time consumption (24ms), further demonstrating the efficiency of our proposed modules.

	We also observe that adding SPN to the high-precision method can paradoxically lead to a decrease in accuracy, which is attributable to network redundancy. This is also true for some complex networks, such as LRRU \cite{wang2023lrru}, indicating that spatial propagation networks are not a panacea in the field of depth completion. Ray Tune's hyperparameter tuning and search for the number of module iterations offer a more cost-effective solution.
	
	\subsection{Pareto Optimality}
	
	To verify that the growth in model accuracy is not solely dependent on the increase in parameters, we analyze the speed and accuracy, transforming the task into a multi-objective optimization process. The objective functions $f_1(x)$ and $f_2(x)$ represent time and RMSE values, respectively. The smaller both are, the better the method. The multi-objective optimization model is defined as follows:
	\begin{equation}
	\left\{
	\begin{aligned}
	& \min_{x \in X} F(x) = [f_1(x), f_2(x)]^T \\
	& \quad X \subseteq \text{KITTI}
	\end{aligned}
	\right.
	\end{equation}
	
	We use Pareto dominance to evaluate any two depth completion methods $a$ and $b$, where $a$ Pareto dominates $b$, denoted as $a > b$, if and only if:
	\begin{equation}
	\left\{
	\forall i \in \{1, 2\}, f_i(a) \leq f_i(b)
	\right\}
	\wedge
	\left\{
	\exists j \in \{1, 2\}, f_j(a) < f_j(b)
	\right\}
	\end{equation}
	
	Let $S$ be the feasible region for the multi-objective optimization, and let $F(x)$ be the vector objective function. For every $x$ in $S$, if $F(x) \leq F(\overline{x})$, then $\overline{x}$ is an efficient solution to the multi-objective optimization task, also known as a non-dominated solution and Pareto optimal solution.
	
\begin{figure}[t]
		\begin{center}
			\includegraphics[width=0.98\linewidth]{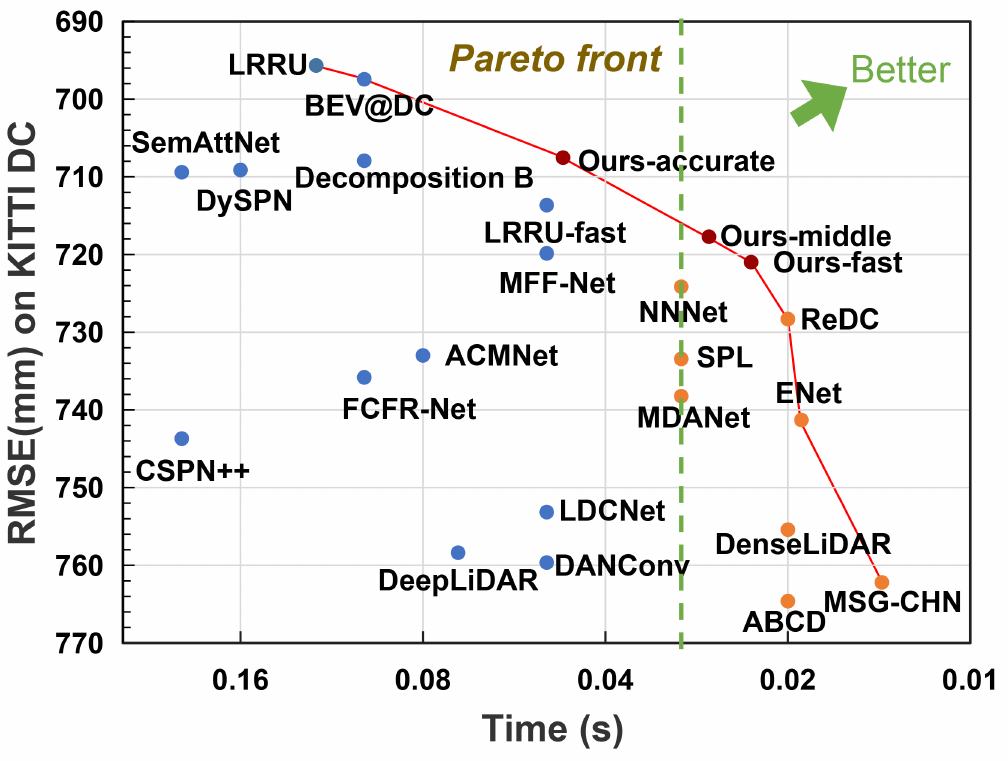}
		\end{center}
		\caption{\textbf{Statistics on speed and accuracy for current depth completion methods, along with the Pareto frontier chart.} The data comes from the official KITTI DC benchmark site. All of our methods have achieved time-accuracy Pareto optimality.}
		\label{fig6}
	\end{figure}

	For a multi-objective optimization task, there are often multiple non-dominated solutions, all of which constitute a non-inferior solution set, also known as the Pareto front, as shown in Figure~\ref{fig6}.
	
	As illustrated in the figure, there are a total of 8 Pareto optimal solutions for the current depth completion methods. Our method ranks first among the methods under 30ms and 100ms, achieving real-time high-precision results.
	
	\section{Conclusion}
	
	The asymmetry between overly dense color information and sparse depth input is a key challenge in depth completion. To address this issue, we propose a Gated Cross-Attention network that uses a cross-attention mechanism to correct color features and complete depth features, and employs confidence propagation for progressive optimization of depth features. At the same time, global perception is achieved at low resolution through a Transformer variant. For optimizing the number of iterations, we employ a machine learning approach using Ray Tune to search for the optimal hyperparameter combinations. We conduct extensive experiments on both indoor and outdoor scene datasets, and our method not only achieves state-of-the-art performance but also delivers fast and accurate results, reaching  Pareto optimality in all cases with good generalizability. In future work, we plan to design more efficient fusion mechanisms through neural architecture search (NAS) to achieve better depth completion results.
	
	\section*{Acknowledgment}
	
	The authors would like to thank all the anonymous reviewers for their insightful reviews and valuable comments on the manuscript, which have helped to improve the quality of the paper. Thanks Jie Tang for providing the code of GuideNet.
	
	\bibliographystyle{abbrv-doi}
	\bibliography{template}
\end{document}